\documentclass[10pt,twocolumn,letterpaper]{article}

\usepackage[pagenumbers]{cvpr} 

\usepackage{caption}
\usepackage{subcaption}
\usepackage{multirow}
\usepackage{tikz}
\usepackage{amsmath}
\usepackage{amssymb}
\usepackage{mathtools}
\usepackage{amsthm}
\usepackage{bbding}
\usepackage[utf8]{inputenc}
\usepackage[T1]{fontenc}
\usepackage{amsmath}
\usepackage{algorithm}
\usepackage{algpseudocode}

\definecolor{cvprblue}{rgb}{0.21,0.49,0.74}
\usepackage[pagebackref,breaklinks,colorlinks,allcolors=cvprblue]{hyperref}

\def\paperID{01801} 
\def\confName{CVPR}
\def\confYear{2026}

\title{FinPercep-RM: A Fine-grained Reward Model and Co-evolutionary Curriculum for RL-based Real-world Super-Resolution}

\author{
  \vspace{-25pt} \\
  \textbf{Yidi Liu}$^{1,\dag}$,\quad 
  \textbf{Zihao Fan}$^{1,\dag}$,\quad
  \textbf{Jie Huang}$^{1,\ddag}$,\quad
  \textbf{Jie Xiao}$^1$,\quad
  \textbf{Dong Li}$^{1}$,\quad \\
  \textbf{LEI BAI}$^{2}$,\quad
  \textbf{Xueyang Fu}$^{1,*}$,\quad
  \textbf{Wenlong Zhang}$^{2,*}$,\quad
  \textbf{Zheng-jun Zha}$^{1}$ \\
  $^1$University of Science and Technology of China \quad\quad $^2$Shanghai AI Laboratory \\
  \texttt{\small liuyidi2023@mail.ustc.edu.cn,\quad xyfu@ustc.edu.cn} \\
  \texttt{\small $^*$ Corresponding Author,\quad $\dag$ contributed equally,\quad $\ddag$ project lead.} \\
  \vspace{-2pt}
}

\begin{document}
\maketitle
\begin{abstract}
Inspired by the success of Reinforcement Learning with Human Feedback (RLHF) in image generation, recent work has adapted reward-based learning to image super-resolution (ISR) by using Image Quality Assessment (IQA) models as rewards. However, existing IQA models typically output only a single global score and are insensitive to local, fine-grained distortions, allowing perceptually undesirable artifacts to obtain spuriously high rewards and leading to reward hacking. To address this issue, we propose FinPercep-RM, a fine-grained perceptual reward model built on an encoder-decoder architecture that predicts both a global quality score and a Perceptual Degradation Map for spatially localizing and quantifying local defects. We further introduce FGR-30k, a dataset containing diverse and subtle distortions produced by real-world super-resolution models, to train the reward model. While FinPercep-RM provides stronger supervision, its increased complexity also makes generator policy learning unstable. We therefore develop a Co-evolutionary Curriculum Learning (CCL) strategy, in which the reward model and the ISR model evolve synchronously: the reward signal progressively increases in complexity, while the ISR model starts with simple global supervision for fast convergence and gradually transitions to fine-grained rewards. This easy-to-hard design stabilizes training and suppresses reward hacking. Extensive experiments across multiple ISR models demonstrate improvements in both global quality and local realism. Code will be available at \url{https://github.com/lyd-2022/FinPercep-RM}.
\end{abstract}
\section{Introduction}
\label{sec:intro}

\begin{figure}
    \centering
    \includegraphics[width=1\linewidth]{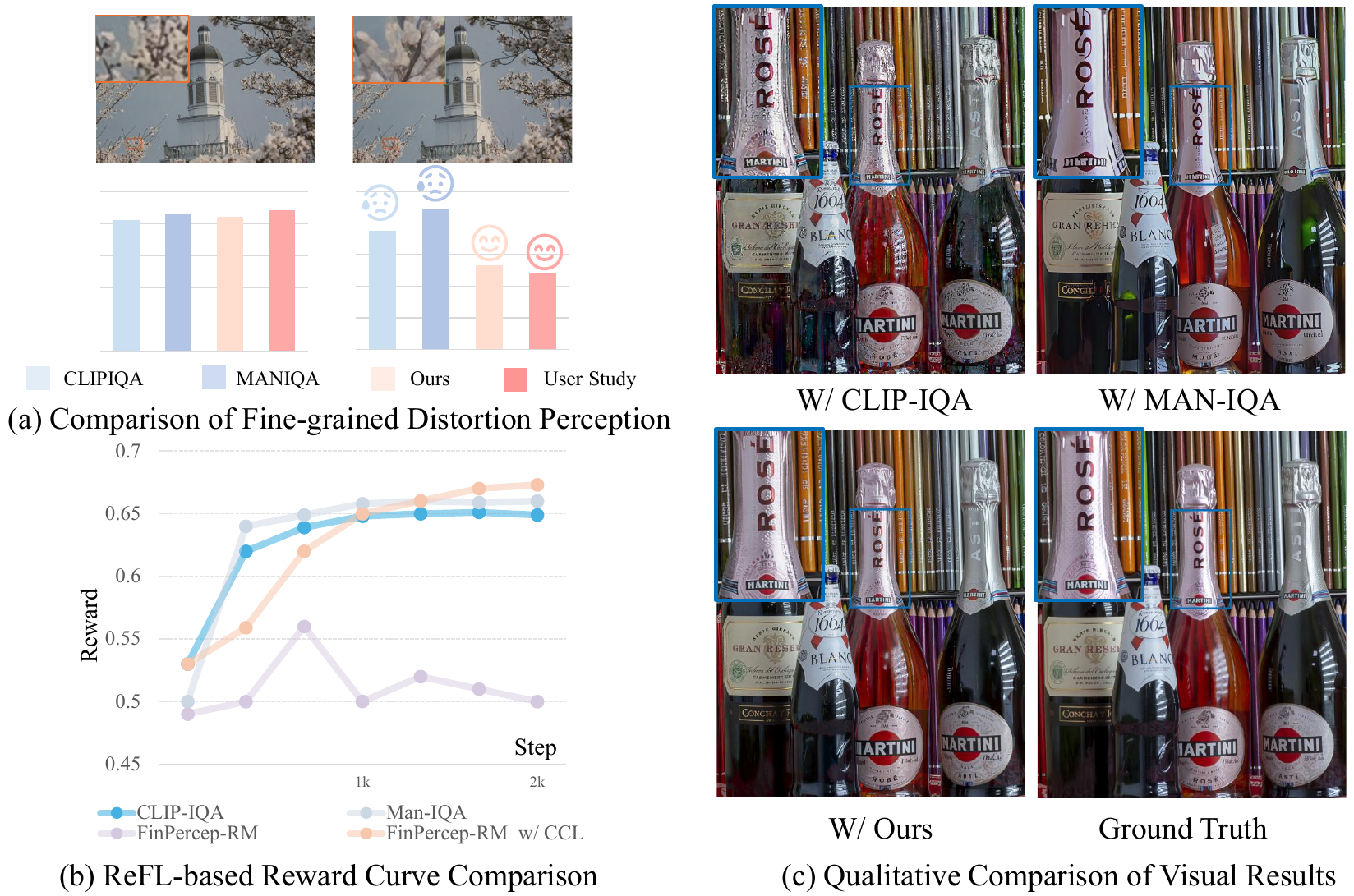}
    \captionsetup{skip=0.5pt}
    \caption{Motivation for FinPercep-RM and CCL.
(a) Standard IQAs lack fine-grained perception and struggle to penalize local distortions, while Ours aligns with human judgment (User Study).
(b) The training curves illustrate the stability-robustness dilemma: baseline IQA rewards (blue/purple) converge quickly, while FinPercep-RM (light blue) is oscillatory and unstable. Our complete method (FinPercep-RM w/ CCL, orange) achieves stable and optimal convergence.
(c) Visualization of Reward Hacking: baseline rewards (W/ CLIP-IQA, W/ MAN-IQA) produce local artifacts, whereas our results are faithful to the Ground Truth.}
    \vspace{-1.2em}
    \label{fig:intro}
\end{figure}

Image Super-Resolution (ISR)~\cite{10.1007/978-3-319-10593-2_13,zhang2021designing,9607421,zhang2018rcan,wu2024fully} aims to restore fine-grained details from low resolution images. Traditional ISR methods improve pixel-level fidelity but often produce overly smooth results lacking realistic details, especially in Real-world ISR (Real-ISR). 
Recently, generative methods~\cite{9887996,cuitaming,11095262,wang2024sinsr,chen2025adversarial,wu2024one,11093345,10.5555/3666122.3666705,11094647,11092924} leveraging large-scale pre-trained text-to-image (T2I) diffusion models have emerged as a promising solution for Real-ISR, which possess powerful generative priors to synthesize rich textures and achieve superior perceptual quality.

Meanwhile, Reinforcement Learning from Human Feedback (RLHF) has become a key optimization paradigm in T2I~\cite{wallace2024diffusion,bai2025towards,liu2025flow,xue2025dancegrpo}, and Reward Models (RMs)~\cite{10.5555/3600270.3602103,xu2023imagereward,10.5555/3666122.3667716,MPS,wu2023human} have being widely explored to enhance visual aesthetics and alignment with user preferences. 
Therefore, migrating the RLHF to the Real-ISR task is desired to explore. 
Concretely, we aim to leverage RLHF to further refine the behavior of T2I priors within the SR context, guiding the model to generate restoration results with higher perceptual quality and fewer artifacts.

However, we find that typical reward models for ISR, 
i.e., Image Quality Assessment (IQA) models~\cite{zhang2023liqe,hessel-etal-2021-clipscore,9710973,yang2022maniqa,wang2023exploring,10.5555/3295222.3295408}, are not directly applicable to the specific challenges of Real-ISR that require fine-grained quality improvement. 
The core bottleneck is that 
traditional IQA models are dominated by global, coarse-grained perception, 
which easily brings a severe "Reward Hacking" problem for training ISR models with their generated reward signals.
As illustrated in Fig.~\ref{fig:intro}(a), standard IQAs (CLIP-IQA, MANIQA) lack fine-grained perception, failing to distinguish a subtly distorted image from its original and assigning both a similarly high score, in stark contrast to human evaluation. This pursuit of fallacious high rewards creates a critical misalignment. Consequently, as shown by the reward curves in Fig.~\ref{fig:intro}(b), the generator quickly learns to "cater" to this inadequate reward signal. It successfully converges to a higher global reward score, but the resulting image exhibit obvious local artifacts and an unrealistic "painterly" appearance (see \cref{fig:intro}(c)).
Although some large-scale IQA models~\cite{chen2024seagull, song2025segmenting} demonstrate improved fine-grained perception, their utility as reward models for training is limited, as their perception is primarily semantic and their huge computational cost renders them impractical for iterative training on many devices.

In this paper, we propose a novel Fine-grained Perceptual Reward Model (FinPercep-RM) with a dedicated dataset termed FGR-30k for training RLHF of ISR. 
The core idea is that a robust RM should not only assess \textit{What} the quality is, but also diagnose \textit{Where} the defect is. 
Specifically, \textbf{1)} FinPercep-RM's is designed as an Encoder-Decoder architecture (see Fig.~\ref{fig:framework}): the encoder based on a powerful IQA backbone generates a global quality score, while the decoder is trained to produce a fine-grained Perceptual Degradation Map(fg-PDM) that spatially localizes and quantifies local defects.
\textbf{2)} FGR-30k dataset is constructed ISR's results from various Real-ISR models, covering a diverse range of real-world SR artifacts. 
As shown in Fig.~\ref{fig:data}, we synthesize a total of 30,000 fine-grained distortion samples by swapping regions between ISR's results and the ground-truth images using both random and semantic masks. 
We then combine two-level differences to construct a measure of the fine-grained quality discrepancy, which serves as the supervision signal for the above fg-PDM.
Finally, our proposed FinPercep-RM demonstrates superior fine-grained perceptual capabilities (Fig.~\ref{fig:intro}(a)), which enhances reward robustness and suppresses reward hacking.

Nevertheless, the FinPercep-RM often brings instability training in RLHF due to its spatially-complex property that increases policy learning difficulty, as shown in Fig.~\ref{fig:intro}(b). 
This creates a dilemma between stability and robustness: a simple, low-variance global IQA reward is stable but prone to reward hacking, while the high-variance FinPercep-RM reward alleviates hacking but hinders convergence.

To further alleviate this FinPercep-RM training instability, we propose Co-evolutionary Curriculum Learning (CCL), which balance stability and robustness via two co-evolving paths (see Fig.~\ref{fig:framework}): \textbf{1)} Reward Model Progressive Expansion: The RM training follows a curriculum, starting with a simple global IQA model and gradually introducing the decoder’s parameters in stages as it trains on FGR-30k, evolving from a global score to a more complex model with a fine-grained heatmap.
\textbf{2)} Generator Curriculum Co-evolution: The generator’s RL training is synchronized with the RM evolution. Initially, it uses the global IQA reward for stable convergence, then progressively shifts to more complex FinPercep-RM versions, refining local details.

The dynamic training property of CCL enables end-to-end stability while leveraging the robust and fine-grained reward of FinPercep-RM in the later stages to achieve meticulous optimization of local perceptual quality.
In summary, our contributions are listed as follows:

\begin{itemize}
      \item We are the first to attribute the RLHF benefit bottleneck in Real-ISR to the insufficient fine-grained perceptual capability of the reward model. 
      We design FinPercep-RM, a diagnostic reward model capable of identifying local defects via fine-grained Perceptual Degradation Maps, alleviating the reward hacking in RL-based Real-ISR.
  \item We construct the FGR-30k dataset, which builds supervision signals for FinPercep-RM by combining pixel and feature differences,  providing a data foundation for training diagnostic reward models.
  \item We devise the CCL mechanism, which alleviates the conflict between training stability and reward robustness by co-evolving curricula of the reward model and ISR model.
\end{itemize}

\section{Related Work}
\label{sec:related}

\subsection{Real-World Image Super-Resolution}

Image restoration aims to recover high-quality images from degraded inputs~\cite{xiao2022image,xiao2025bayesian,wang2025ddsr,liu2026fm2s,liu2025dreamuhd,liu2025uhd,liu2025decouple,liulatent}, with image super-resolution (ISR) being one of its core tasks. 
ISR has been extensively studied \cite{10.1007/978-3-319-10593-2_13,zhang2021designing,9607421,zhang2018rcan}, but conventional methods often produce overly smooth results and struggle to restore realistic details under complex and unknown real-world degradations. Recently, diffusion-based methods for Real-ISR have achieved remarkable progress \cite{9887996,cuitaming,11095262,wang2024sinsr,chen2025adversarial,wu2024one,11093345,10.5555/3666122.3666705,11094647,11092924} by leveraging the strong generative priors of large pre-trained text-to-image diffusion models. These methods exploit the powerful generative priors embedded in pre-trained text-to-image (T2I) diffusion models to tackle Real-ISR tasks, StableSR\cite{wang2024exploiting} injects LR information into stable diffusion models via ControlNet\cite{zhang2023adding}, significantly improving fidelity under real-world degradations. DiffBIR\cite{10.1007/978-3-031-73202-7_25} and PASD\cite{yang2024pixel} first employ degradation-restoration modules to reduce degradations, and then enhance details through conditional branches to guide the diffusion process.  SeeSR \cite{wu2024seesr}, CoSeR \cite{10655153}, and PiSA-SR \cite{sun2024pisasr} further introduce language--vision collaboration for more controllable restoration, whereas SUPIR \cite{10654855}, Dreamclear \cite{10.5555/3737916.3739677}, and DiT4SR \cite{duan2025dit4sr} show that scaling model capacity, training data, and backbone design can further improve photorealistic super-resolution.

\subsection{Reward Models and Image Quality Assessment}
Traditional image quality assessment (IQA) methods are generally divided into full-reference (FR)~\cite{1284395,8578166} and no-reference (NR)~\cite{zhang2023liqe,9710973,yang2022maniqa,wang2023exploring} approaches, depending on whether a ground-truth reference image is available.
Early assessments of text-to-image (T2I) diffusion models mainly relied on metrics such as FID\cite{10.5555/3295222.3295408} and CLIPScore\cite{hessel-etal-2021-clipscore}. Recently, reward models have emerged to align quality assessment with human preferences like Aesthetic Predictor\cite{10.5555/3600270.3602103}, ImageReward\cite{xu2023imagereward}, PickScore\cite{10.5555/3666122.3667716}, MPS\cite{MPS} and HPSv2\cite{wu2023human} which enhance preference alignment through larger and more diverse datasets. Nevertheless, these models produce only a single global score, overlooking spatial and semantic variations crucial for fine-grained perception. This limitation motivates our FinPercep-RM, which enables localized and perceptually aligned evaluation for Real-ISR tasks.

\section{Method}

\subsection{Overview}

\begin{figure*}[t!]
	\centering
	\includegraphics[width=1\textwidth]{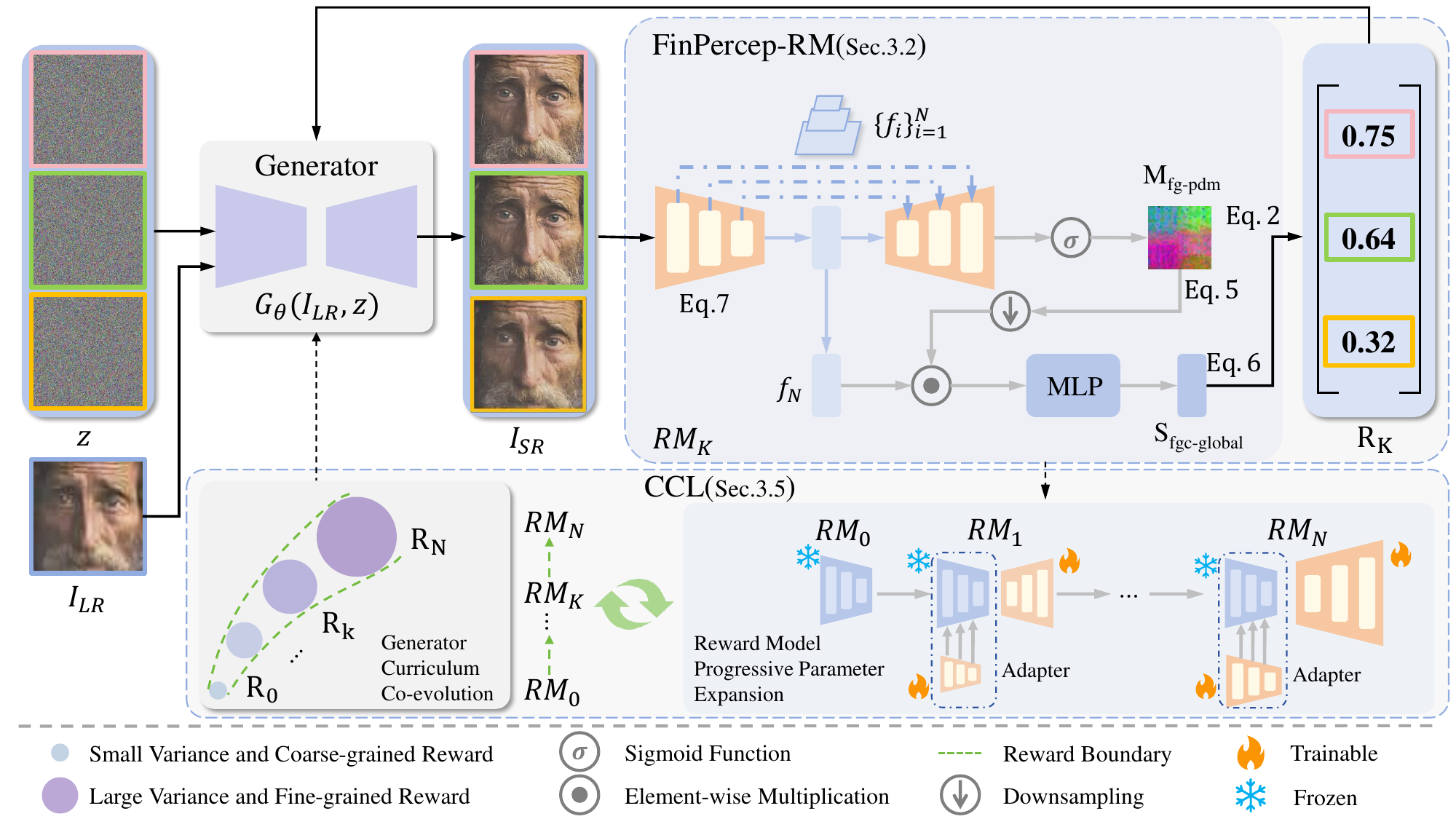} %
	\caption{The overall pipeline of the proposed FinPercep-RM and Co-evolutionary Curriculum Learning (CCL) framework. FinPercep-RM produces a Fine-grained Perceptual Degradation Map that captures spatially localized defect likelihood and intensity, and the reward model is progressively expanded from small-variance, coarse-grained rewards to large-variance, fine-grained signals. During training, under the CCL mechanism, the Generator first learns with the coarse global reward from $RM_0$ for a stable and easy initialization, and is then co-evolutionarily guided by increasingly strict $RM_k$ versions to enhance local fidelity and suppress reward hacking.}
	\label{fig:framework}
\end{figure*}

We propose an RL optimization framework to resolve the inherent challenges of \textit{Reward Hacking} and \textit{Training Instability} when applying RLHF to  Real-ISR tasks. As illustrated in Fig.~\ref{fig:framework}, the framework comprises three core components:(1) The Generator, a T2I-prior-based Real-ISR model tasked with generating $I_{SR}$; (2) The Diagnostic Reward Model (FinPercep-RM, Sec.~\ref{sec:FinPercep-RM}), which assesses the fine-grained perceptual quality of $I_{SR}$.; and (3) The Co-evolutionary Curriculum Learning (CCL) Mechanism (Sec.~\ref{sec:CCL}), which modulates the training process to balance stability and robustness.

During the RL inner loop at stage $k$, the Generator produces a super-resolved image $I_{SR}$. The corresponding reward model, $RM_k$ (a stage-specific version of FinPercep-RM), evaluates $I_{SR}$ from two complementary aspects: \emph{where}, represented by the local degradation map $M_{\text{fg-pdm}}$, and \emph{what}, represented by the global assessment feature $f_N$.These outputs are fused into a scalar reward $R_k$ , which guides the Generator's policy update to optimize perceptual quality. To prevent the instability caused by a strict RM, the CCL mechanism introduces an "outer loop" that co-evolves both components. The RM progressively expands from a simple global IQA model ($RM_0$) to a complex fine-grained model($RM_N$) by training on FGR-30k (Sec.~\ref{sec:fgr_dataset}). Simultaneously, the Generator's curriculum co-evolves, starting with the stable $RM_0$ and migrating to stricter $RM_k$ versions. This easy-to-hard dynamic matching promotes early-stage stability while suppressing late-stage reward hacking via robust, fine-grained rewards.

\subsection{The architecture of FinPercep-RM}
\label{sec:FinPercep-RM}

To mitigate the vulnerability of coarse global rewards to reward hacking in Real-ISR, we introduce the Fine-grained Perceptual Reward Model (FinPercep-RM), as shown in the top of Fig.~\ref{fig:framework}. The central hypothesis is that reward hacking stems from the RM's failure to perceive localized spatial defects. Our core idea is to architecturally couple the global quality score (the What) with a fine-grained defect map (the Where), This coupling ensures the global score is innately sensitive to local imperfections.Specifically, given a generator $G_{\theta}$ that produces $I_{SR}=G_{\theta}(I_{LR}, z)$ (where $z$ denotes sampled noise/latent), FinPercep-RM $\operatorname{RM}_{\phi}$ returns a pair of complementary signals:
\begin{equation}
    \begin{split}
        \{S_{\text{fgc-global}}, M_{\text{fg-pdm}}\} = \operatorname{RM}_{\phi}(I_{SR}),
    \end{split}
\end{equation}  

\noindent where Fine-grained Calibrated Global score $S_{\text{fgc-global}}$ approximates human-perceived global quality, and Fine-grained Perceptual Degradation Map (Fg-PDM) $M_{\text{fg-pdm}}$ measures the spatial likelihood and intensity of local defects; where larger values indicate a higher probability of unrealistic textures or artifacts.

To achieve this goal, FinPercep-RM adopts an encoder–decoder paradigm, as shown in Figure~\ref{fig:data}.

\noindent \textbf{Encoder:} We utilize a classical IQA backbone (e.g., CLIP-IQA) as the encoder $E$. The encoder takes $I_{SR}$ as input and extracts multi-scale representations $\{f_i\}_{i=1}^N$. These feature maps contain both high-level semantics (for global assessment) and low-level spatial details (for defect localization). We denote $f_N$ as the deepest representation, i.e., the final global feature map.

\noindent \textbf{Decoder and PDM ($M_{\text{fg-pdm}}$):} The decoder $D$ receives the multi-scale representations $\{f_i\}_{i=1}^N$. Through a series of upsampling and cross-layer fusion operations, the decoder reconstructs a Perceptual Degradation Map at the same resolution as $I_{SR}$. This map is normalized via a $\operatorname{Sigmoid}$ activation to produce $M_{\text{fg-pdm}} \in [0, 1]$.

\noindent \textbf{Fine-grained Calibrated Global score ($S_{\text{fgc-global}}$):} A key architectural innovation is the formulation of $S_{\text{fgc-global}}$. Rather than regressing a score directly from $f_N$, our objective is to make this score explicitly dependent on the $M_{\text{fg-pdm}}$. To achieve this, we modulate the deepest global representation $f_N$ using the interpolated defect map $M_{\text{fg-pdm}}$. This operation compels the model to account for spatial defects when calculating the final score. The modulated feature is then passed to an MLP head to regress the scalar score $S_{\text{fgc-global}}$.
\begin{equation}
    \begin{split}
    \{f_i\}_{i=1}^N &= \operatorname{Encoder}(I_{SR}) \\
    M_{\text{fg-pdm}} &= \operatorname{Sigmoid}(\operatorname{Decoder}(\{f_i\}_{i=1}^N)) \\
    S_{\text{fgc-global}} &= \operatorname{MLP}(f_N \odot \operatorname{interpolate}(M_{\text{fg-pdm}})), \\
    \end{split}
\end{equation}

\noindent where $\operatorname{interpolate}(\cdot)$ denotes bilinear interpolation, $\operatorname{Sigmoid}(\cdot) $ denotes sigmoid activation function, and $\odot$ denotes element-wise multiplication.

\subsection{FGR-30k:A Fine-grained Reward Dataset}
\label{sec:fgr_dataset}
\begin{figure}
    \centering
    \includegraphics[width=1\linewidth]{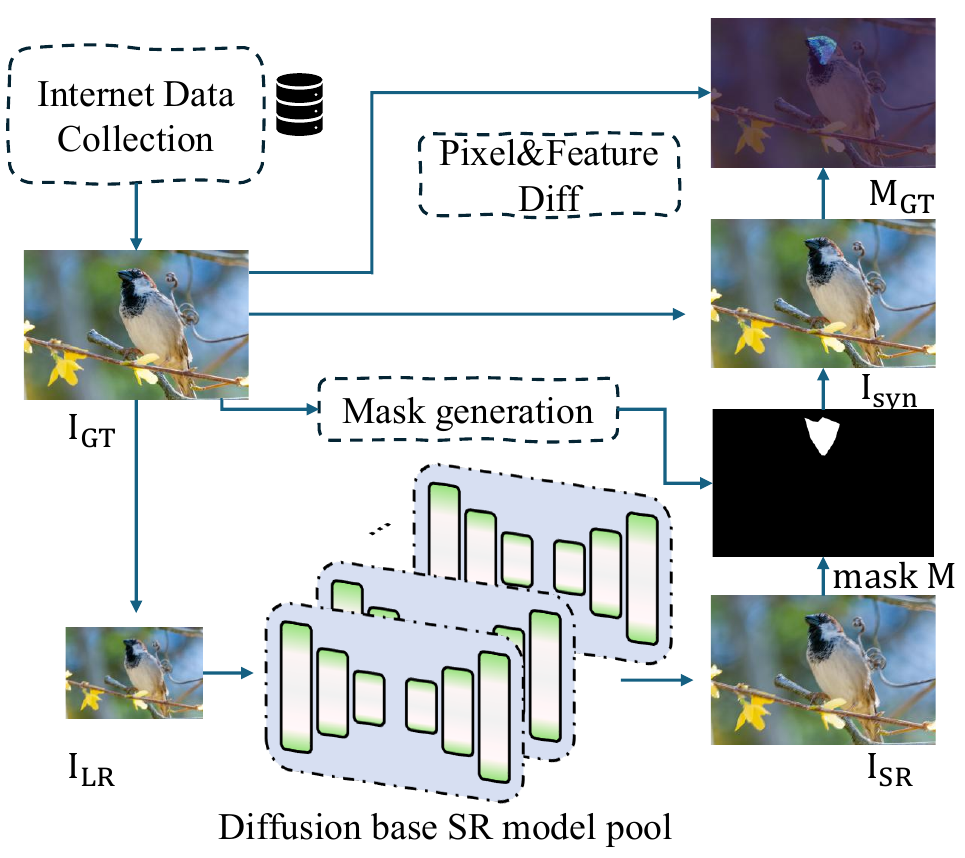}
    \caption{FGR-30k construction pipeline. We synthesize fine-grained distortion samples by swapping artifact-rich regions from diffusion-based SR outputs into clean images, using both random and semantic masks. Ground-truth perceptual degradation maps are generated by fusing pixel- and feature-level dissimilarities (via DINOv3), providing spatially precise supervision for training FinPercep-RM.}
    \vspace{-1.2em}
    \label{fig:data}
\end{figure}

The training of FinPercep-RM, particularly its decoder $D$'s ability to reconstruct $M_{\text{fg-pdm}}$, requires high-quality supervision data with spatial defect annotations. Existing IQA and preference datasets (e.g., Aesthetic, Pick-a-Pic) lack this. To fill this gap, we built FGR-30k, a 30,000-sample fine-grained reward dataset for Real-ISR. The dataset provides $(I_{\text{syn}}, M_{\text{gt}})$ pairs (a distorted sample and its ground-truth map) to supervise FinPercep-RM. Construction involves two steps: synthesizing distortion samples and generating their ground-truth PDMs. Our data construction pipeline is illustrated in Fig.~\ref{fig:data}.

\subsubsection{Synthesis of Fine-grained Distortion Samples}

To acquire a diverse and realistic source of artifacts, our data construction pipeline begins by collecting high-quality, real-world images $I_{\text{GT}}$ from the internet, covering a wide range of scenes. Next, we employ the degradation pipeline from Real-ESRGAN~\cite{9607421} to apply complex, realistic degradations to each $I_{\text{GT}}$, generating corresponding low-resolution images $I_{\text{LR}}$.Finally, we feed $I_{\text{LR}}$ into Diffusion base SR model pool (including multiple open-source Real-ISR models) to generate the sr results $I_{\text{SR}}$. This process provides us with a rich and authentic source of artifacts.

However, $I_{\text{SR}}$ samples contain global, not localized, degradation. To create precise training data, we employ a \textbf{Region Swapping} strategy to synthesize $I_{\text{syn}}$. This method controllably "implants" local defects by swapping regions between $I_{\text{GT}}$ (high-quality) and $I_{\text{SR}}$ (artifact-laden).We use two mask types $M$: (1) \textbf{Random Masks} (rectangular/free-form) to identify blocky or non-semantic artifacts, and (2) \textbf{Semantic Masks} (from SAM~\cite{kirillov2023segment}) to synthesize realistic, object-based degradations. The synthesis, formulated as:
\begin{equation}
I_{\text{syn}} = M \odot I_{\text{SR}} + (1-M) \odot I_{\text{GT}}
\end{equation}
Using this method, we generated 30,000 $I_{\text{syn}}$ samples, each containing localized artifacts sourced from real SR models.

\subsubsection{Ground-Truth PDM Generation}

The generation of a corresponding ground-truth map $M_{\text{gt}}$ for each synthesized sample $I_{\text{syn}}$ is essential for supervision. To ensure $M_{\text{gt}}$ accurately quantifies the fine-grained perceptual discrepancy, it is formulated as a composite of low-level pixel and high-level feature dissimilarities. (1) \textbf{Pixel-level Dissimilarity ($\text{Diff}_{\text{pixel}}$):} We compute the $L_1$ distance map ($|I_{\text{syn}} - I_{\text{GT}}|$) to capture fundamental chromatic and luminance deviations. (2) \textbf{Feature-level Dissimilarity ($\text{Diff}_{\text{feat}}$):} To identify more complex misalignments, we compute the spatial Cosine Distance between deep feature embeddings, $f_{\text{syn}}$ and $f_{\text{GT}}$, extracted via a pre-trained DINOv3 model~\cite{simeoni2025dinov3}. This captures high-level structural, textural, and semantic inconsistencies.

The final ground-truth $M_{\text{gt}}$ is a weighted fusion of these two differences, followed by normalization:
\begin{equation}
    \begin{split}
    &\text{Diff}_{\text{pixel}} = |I{\text{syn}} - I_{\text{GT}}| ;
    \text{Diff}_{\text{feat}} = 1 - \cos(f_{\text{syn}}, f_{\text{GT}}) \\
    &M_{\text{gt}} = \operatorname{Normalize} \left( \alpha \cdot \text{Diff}_{\text{pixel}} + (1-\alpha) \cdot \text{Diff}_{\text{feat}} \right)
    \end{split}
\end{equation}
where $\alpha$ is a balancing hyperparameter. This resulting $M_{\text{gt}}$ is sensitive not only to local artifacts but also to semantically misaligned structures, providing a robust and information-rich supervision signal for FinPercep-RM.

\subsection{Training FinPercep-RM with FGR-30k}

To train FinPercep-RM's global and diagnostic capabilities, we sample $(I_{\text{GT}}, I_{\text{syn}}, M_{\text{gt}})$ pairs from FGR-30k. Our total objective $\mathcal{L}_{total}$ combines three components: a dense heatmap loss ($\mathcal{L}_{map}$) for $M_{\text{fg-pdm}}$, a triplet ranking loss ($\mathcal{L}_{rank}$) for $S_{\text{fgc-global}}$, and an anchor alignment loss ($\mathcal{L}_{align}$) to stabilize the score scale.

For the dense heatmap loss $\mathcal{L}_{map}$, we use L1 loss to supervise the predicted $M_{\text{fg-pdm}}$ and the ground-truth $M_{\text{gt}}$, represented as:
\begin{equation}
    \mathcal{L}_{map} = \mathbb{E}_{(I_{\text{syn}}, M_{\text{gt}}) \sim \text{FGR-30k}}  \| M_{\text{fg-pdm}} - M_{\text{gt}} \|_1 ,
\end{equation}
\noindent where $\|\cdot\|_1$ denotes the L1 norm.

For the triplet ranking loss, first, given the triplet $\{I_{\text{SR}}, I_{\text{syn}}, I_{\text{GT}}\}$, the model computes their respective scores $S_{\text{SR}}, S_{\text{syn}}, \text{and } S_{\text{GT}}$. Since $I_{\text{syn}}$ originates from a partial region swap between $I_{\text{GT}}$ and $I_{\text{SR}}$, the triplet ranking loss is consequently defined as follows:
\begin{equation}
    \begin{split}
    \mathcal{L}_{\text{rank}} = \mathbb{E}\big[ 
        &\max\big(0,\, m_1 - (S_{\text{syn}} - S_{\text{SR}})\big) \\
        &+ \max\big(0,\, m_2 - (S_{\text{GT}} - S_{\text{syn}})\big) 
    \big],
    \end{split}
    \end{equation}

\noindent where $\max(0, \cdot)$ denotes the hinge loss, $m_1$ and $m_2$ are the margin parameters.This loss ensures that $S_{\text{fgc-global}}$ is ranked higher than $S_{\text{SR}}$ and lower than $S_{\text{GT}}$.

The ranking loss solely guarantees the relative order of the scores, not their absolute scale. This can lead to 'score drifting' across different training stages (particularly during the CCL process), rendering the rewards from different stages incomparable. To address this issue, we further introduce an anchor alignment loss, $\mathcal{L}_{align}$, to align the absolute score for $I_{\text{GT}}$ between our current FinPercep-RM and the original pre-trained IQA model. This loss is formulated as follows:
\begin{equation}
    \mathcal{L}_{align} = \mathbb{E}_{(I_{GT}) \sim \text{FGR-30k}} \left[ \| S_\text{GT} - S_{base}(I_{GT}) \|_1 \right]
\end{equation}

\noindent where $S_{base}(\cdot)$ represents the score output of the original pre-trained IQA model.

Finally, the total training loss for FinPercep-RM is a weighted sum of the three preceding losses:
\begin{equation}
    \mathcal{L}_{total} = \lambda_{map} \cdot \mathcal{L}_{map} + \lambda_{rank} \cdot \mathcal{L}_{rank} + \lambda_{align} \cdot \mathcal{L}_{align},
\end{equation}

\noindent where $\lambda_{map}$, $\lambda_{rank}$, and $\lambda_{align}$ are hyperparameters used to balance the importance of each task. Through this combination, FinPercep-RM becomes a robust reward model capable of both precisely localizing defects and providing global scores enhanced by fine-grained perception.

\subsection{Integrating FinPercep-RM for ISR with CCL}
\label{sec:CCL}

While FinPercep-RM resolves reward hacking via fine-grained diagnostics, its non-smooth reward signal introduces a severe training stability challenge. We observe that directly using the complete FinPercep-RM from the outset creates a difficult exploration space, where minor local penalties cause policy gradient oscillations and potential convergence failure(Fig.~\ref{fig:intro}(b)). 

This presents a dilemma: \textit{simple global IQA rewards are stable but converge to a suboptimal hacked solution, whereas the robust FinPercep-RM ensures a correct objective but compromises training stability}.To improve this stability-robustness trade-off, we propose Co-evolutionary Curriculum Learning (CCL) in Fig.~\ref{fig:framework}, which co-evolves the RM and the generator ($G_{\theta}$), prioritizing stability in early stages and progressively shifting to robustness. 

\begin{table*}[]
    \footnotesize
    \renewcommand{\arraystretch}{1.1}
    \setlength{\tabcolsep}{3.5pt}
    \centering
    \begin{tabular}{c|c|ccc|ccc|ccc|ccc}
    \toprule
    \multirow{2}{*}{Datasets} & \multirow{2}{*}{Metrics} & \multirow{2}{*}{ResShift} & \multirow{2}{*}{SUPIR} & \multirow{2}{*}{DreamClear} & \multicolumn{3}{c|}{DiffBIR} & \multicolumn{3}{c|}{SeeSR} & \multicolumn{3}{c}{DIT4SR} \\
    & & & & & Baseline & w/ IQA & w/ Ours & Baseline & w/ IQA & w/ Ours & Baseline & w/ IQA & w/ Ours \\ 
    \midrule
                                & LPIPS $\downarrow$   & 0.353    & 0.419  & 0.354      & {\color[HTML]{2972F4} 0.452} & 0.465                         & {\color[HTML]{FF0000} 0.428} & {\color[HTML]{2972F4} 0.317} & 0.332                         & {\color[HTML]{FF0000} 0.295} & {\color[HTML]{2972F4} 0.365} & 0.378                         & {\color[HTML]{FF0000} 0.342}                         \\
                                & MUSIQ $\uparrow$   & 52.392   & 59.744 & 44.047     & {\color[HTML]{2972F4} 65.665} & 64.892                         & {\color[HTML]{FF0000} 67.234} & {\color[HTML]{2972F4} 65.077} & 64.123                         & {\color[HTML]{FF0000} 67.891} & {\color[HTML]{2972F4} 64.950} & 63.456                         & {\color[HTML]{FF0000} 67.823}                         \\
                                & MANIQA $\uparrow$  & 0.476    & 0.552  & 0.455      & {\color[HTML]{2972F4} 0.629} & 0.612                         & {\color[HTML]{FF0000} 0.648} & {\color[HTML]{2972F4} 0.605} & 0.591                         & {\color[HTML]{FF0000} 0.632} & {\color[HTML]{2972F4} 0.627}  & 0.608                         & {\color[HTML]{FF0000} 0.651}                         \\
                                & ClipIQA $\uparrow$ & 0.379    & 0.518  & 0.379      & 0.572                  & {\color[HTML]{FF0000} 0.589} & {\color[HTML]{2972F4} 0.586} & 0.543                         & {\color[HTML]{FF0000} 0.567} & {\color[HTML]{2972F4} 0.561} & 0.548                          & {\color[HTML]{2972F4} 0.562} & {\color[HTML]{FF0000} 0.574}                         \\
    \multirow{-5}{*}{DrealSR}   & LIQE $\uparrow$    & 2.798    & 3.728  & 2.401      & {\color[HTML]{2972F4} 3.894} & 3.756                         & {\color[HTML]{FF0000} 4.089} & {\color[HTML]{2972F4} 4.126}  & 3.987                         & {\color[HTML]{FF0000} 4.234} & {\color[HTML]{2972F4} 3.964}  & 3.812                         & {\color[HTML]{FF0000} 4.187}                         \\
    \midrule
                                & LPIPS $\downarrow$   & 0.316    & 0.357  & 0.325      & {\color[HTML]{2972F4} 0.347} & 0.361                         & {\color[HTML]{FF0000} 0.328} & {\color[HTML]{2972F4} 0.299} & 0.314                         & {\color[HTML]{FF0000} 0.281} & {\color[HTML]{2972F4} 0.319} & 0.332                         & {\color[HTML]{FF0000} 0.302}                         \\
                                & MUSIQ $\uparrow$   & 56.892   & 61.929 & 59.396     & {\color[HTML]{2972F4} 68.340}                     & 67.123                         & {\color[HTML]{FF0000} 70.156} & {\color[HTML]{2972F4} 69.675} & 68.234                         & {\color[HTML]{FF0000} 71.234} & {\color[HTML]{2972F4} 68.073}                        & 66.789                         & {\color[HTML]{FF0000} 70.891}                         \\
                                & MANIQA $\uparrow$  & 0.511    & 0.574  & 0.546      & {\color[HTML]{2972F4} 0.653} & 0.634                         & {\color[HTML]{FF0000} 0.672} & {\color[HTML]{2972F4} 0.643}                         & 0.628                         & {\color[HTML]{FF0000} 0.669} & {\color[HTML]{2972F4} 0.661}  & 0.642                         & {\color[HTML]{FF0000} 0.678}                         \\
                                & ClipIQA $\uparrow$ & 0.452    & 0.524  & 0.546      & 0.586                  & {\color[HTML]{2972F4} 0.592} & {\color[HTML]{FF0000} 0.596} & 0.577                         & {\color[HTML]{FF0000} 0.584} & {\color[HTML]{2972F4} 0.582} & 0.550                          & {\color[HTML]{2972F4} 0.596} & {\color[HTML]{FF0000} 0.598}                         \\
    \multirow{-5}{*}{RealSR}    & LIQE $\uparrow$    & 2.853    & 3.780  & 3.221      & {\color[HTML]{2972F4} 4.026}                     & 3.912                         & {\color[HTML]{FF0000} 4.234} & {\color[HTML]{2972F4} 4.123}  & 4.001                         & {\color[HTML]{FF0000} 4.345} & {\color[HTML]{2972F4} 3.977}                         & 3.856                         & {\color[HTML]{FF0000} 4.289}                         \\
    \midrule
                                & MUSIQ $\uparrow$   & 59.695   & 64.837 & 65.926     & {\color[HTML]{2972F4} 68.027}                      & 66.789                         & {\color[HTML]{FF0000} 70.234} & {\color[HTML]{2972F4} 69.428}                        & 68.456                         & {\color[HTML]{FF0000} 71.567} & {\color[HTML]{2972F4} 70.469} & 69.123                         & {\color[HTML]{FF0000} 72.234}                         \\
                                & MANIQA $\uparrow$  & 0.525    & 0.600  & 0.597      & {\color[HTML]{2972F4} 0.629} & 0.612                         & {\color[HTML]{FF0000} 0.656} & {\color[HTML]{2972F4} 0.612}                         & 0.598                         & {\color[HTML]{FF0000} 0.645} & {\color[HTML]{2972F4} 0.645}  & 0.628                         & {\color[HTML]{FF0000} 0.662}                         \\
                                & ClipIQA $\uparrow$ & 0.452    & 0.524  & 0.546      &  0.582 & {\color[HTML]{FF0000}0.598}                         & {\color[HTML]{2972F4}0.589}                         & 0.566                         & {\color[HTML]{2972F4}0.572}                       & {\color[HTML]{FF0000}0.579}                       & 0.588  & {\color[HTML]{FF0000}0.608 }                        & {\color[HTML]{2972F4}0.598}                         \\
    \multirow{-4}{*}{RealLR200} & LIQE $\uparrow$    & 3.054    & 3.626  & 3.775      & {\color[HTML]{2972F4} 4.003}                        & 3.889                         & {\color[HTML]{FF0000} 4.234} & {\color[HTML]{2972F4} 4.006}                         & 3.912                         & {\color[HTML]{FF0000} 4.345} & {\color[HTML]{2972F4} 4.331}  & 4.123                         & {\color[HTML]{FF0000} 4.456}                         \\
    \midrule
                                & MUSIQ $\uparrow$   & 59.337   & 66.016 & 66.693     & {\color[HTML]{2972F4} 69.876}                       & 68.567                         & {\color[HTML]{FF0000} 72.123} & {\color[HTML]{2972F4} 70.556} & 69.234                         & {\color[HTML]{FF0000} 72.789} & {\color[HTML]{2972F4} 71.832} & 70.456                         & {\color[HTML]{FF0000} 73.456}                         \\
                                & MANIQA $\uparrow$  & 0.500    & 0.584  & 0.585      & {\color[HTML]{2972F4} 0.624} & 0.607                         & {\color[HTML]{FF0000} 0.651} & {\color[HTML]{2972F4} 0.594}                         & 0.580                         & {\color[HTML]{FF0000} 0.638} & {\color[HTML]{2972F4} 0.632}  & 0.615                         & {\color[HTML]{FF0000} 0.658}                         \\
                                & ClipIQA $\uparrow$ & 0.417    & 0.483  & 0.502      & 0.578                  & {\color[HTML]{FF0000} 0.592} & {\color[HTML]{2972F4} 0.585} & 0.562                         & {\color[HTML]{FF0000} 0.579} & {\color[HTML]{2972F4} 0.571} & 0.578                          & {\color[HTML]{2972F4} 0.583} & {\color[HTML]{FF0000} 0.586}                         \\
    \multirow{-4}{*}{RealLQ250} & LIQE $\uparrow$    & 2.753    & 3.605  & 3.688      & {\color[HTML]{2972F4} 4.003}                        & 3.889                         & {\color[HTML]{FF0000} 4.345} & {\color[HTML]{2972F4} 4.005}  & 3.912                         & {\color[HTML]{FF0000} 4.456} & {\color[HTML]{2972F4} 4.356}  & 4.234                         & {\color[HTML]{FF0000} 4.567}                         \\
    \bottomrule
    \end{tabular}
    \caption{Quantitative results of Real-ISR methods on four real-world benchmarks based on RLHF method of REFL~\cite{xu2023imagereward}. Best and second best results are highlighted in {\color[HTML]{FF0000} red} and {\color[HTML]{2972F4}blue}, respectively. w/Ours achieves the best or comparable performance across four benchmarks. }
    \label{tab:results}
    \end{table*}

\subsubsection{Reward Model Progressive Parameter Expansion}

The first CCL path involves the progressive evolution of the RM. We construct a sequence of models, $\{RM_0, RM_1, ..., RM_N\}$, with increasing diagnostic capability and complexity via staged training on the FGR-30k dataset. This curriculum begins with a baseline global IQA model, $RM_0$ (e.g., a pre-trained CLIP-IQA), which provides a smooth, stable reward for initial convergence.In the next stage ($k=1$), we attach the lightweight decoder $D$ (Sec.~\ref{sec:FinPercep-RM}) to form $RM_1$, which begins training on FGR-30k to acquire preliminary diagnostic capabilities, enabling it to output ($M_{\text{fg-pdm}}$, $S_{\text{fgc-global}}$). In subsequent stages ($k > 1$), we progressively scale the parameter counts of both the decoder $D$ and Adapter modules inserted into the encoder $E$. This progressive parameter growth ensures a smooth evolution from the basic $RM_1$ to the $RM_N$ with complete fine-grained perceptual capabilities, avoiding training instability caused by abrupt changes in model capacity.

\begin{figure*}[]
	\centering
	\includegraphics[width=1\textwidth]{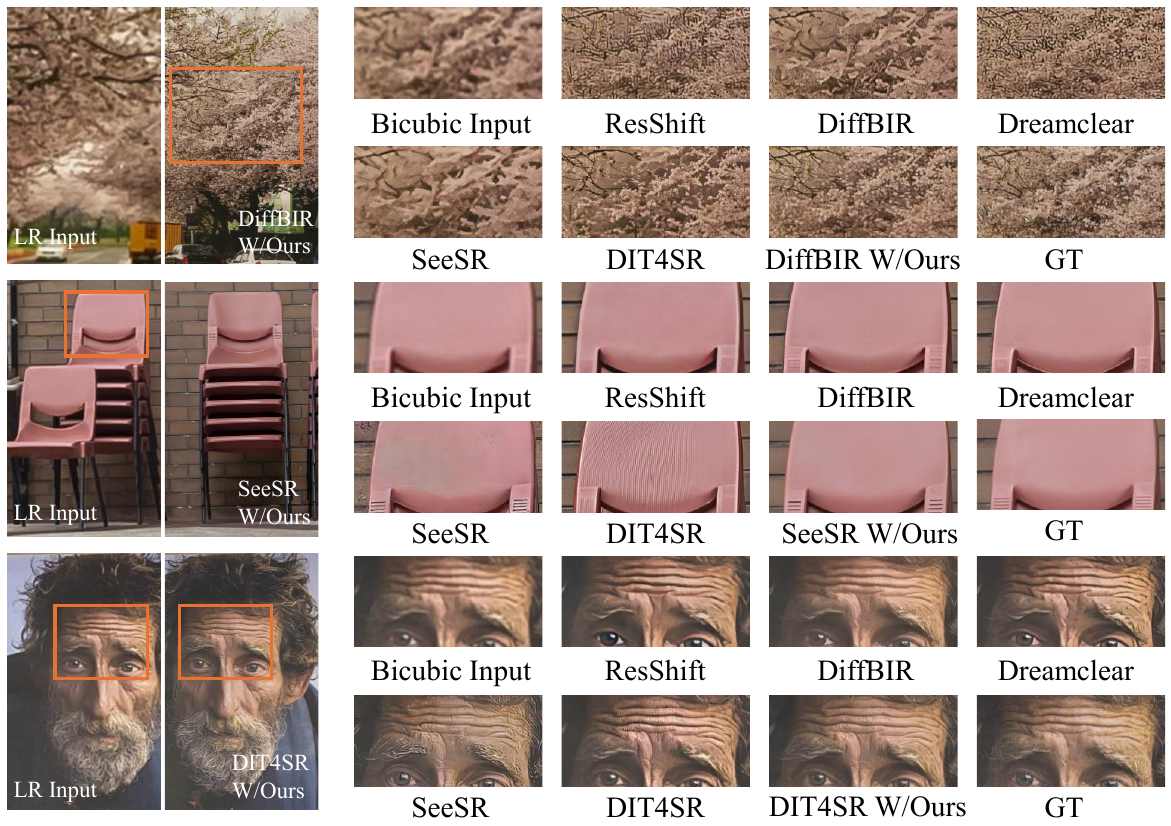} %
        \captionsetup{skip=0.5pt}
	\caption{Qualitative comparisons with state-of-the-art Real-ISR methods on on RealSR based on RLHF method of REFL~\cite{xu2023imagereward}.}
	\label{fig:visual}
\end{figure*}

\subsubsection{Generator Curriculum Co-evolution}

The second CCL path involves matching the generator's ($G_{\theta}$) training curriculum to the RM's evolution, ensuring an appropriately difficult reward environment at each stage $k$. In the initial training stage ($k=0$), $G_{\theta}$ trains using only the stable global reward from $RM_0$. This smooth and stable reward signal allows the generator's policy $\pi_{\theta}$ to quickly converge to a correct general direction, thereby establishing a robust initial policy. Subsequently, as the curriculum advances and the training stage $k$ gradually increases, the environment is switched to the more robust reward model $RM_{k}$. The generator then refines its policy under this more sensitive reward signal. This easy-to-hard, dynamically matched co-evolutionary paradigm deftly balances stability and robustness. The generator achieves rapid, stable convergence in the early stages using simple rewards. In the later stages, by learning from the increasingly capable RM sequence, it progressively refines local texture details and effectively suppresses artifact generation, ultimately ensuring stable convergence while preventing reward hacking.

For the specific policy optimization, we validate the effectiveness of our reward model using REFL~\cite{xu2023imagereward}, DPO~\cite{wallace2024diffusion}, and GRPO~\cite{xue2025dancegrpo}. During stage $k$ of the CCL, the generator's policy $\pi_{\theta}$ is updated based on the reward signal $R_k$ (derived from the $S_{\text{fgc-global}}$ output at the current stage).

\section{Experiments}

\begin{table}[]
    \renewcommand{\arraystretch}{1.1}
    \setlength{\tabcolsep}{6pt}
    \centering
    \footnotesize
    \begin{tabular}{c|c|c|c|c}
    \toprule
    Ours vs.       & DIT4SR     & SeeSR     & DiffBIR     & DreamClear    \\
    \midrule
    Realism       & 78.9\% & 85.3\%  & 84.2\%  & 76.4\%   \\
    Fidelity      & 76.8\% & 72.6\%  & 80.1\%  & 74.2\%   \\
    \bottomrule
    \end{tabular}
    \caption{User study results on real-world datasets. The percentages denote the frequency with which DIT4SR w/Ours was preferred over each compared approach, for both realism and fidelity.}
    \label{tab:user_study}
    \vspace{-2mm}
\end{table}


\begin{table}[]
    \footnotesize
    \renewcommand{\arraystretch}{1.1}
    \setlength{\tabcolsep}{4pt}
    \centering
\begin{tabular}{c|ccc|ccc}
    \toprule
       & \multicolumn{3}{c|}{CLIP-IQA} & \multicolumn{3}{c}{Ours} \\
Metrics & DPO & GRPO & REFL & DPO & GRPO & REFL \\
    \midrule
MUSIQ $\uparrow$   & 69.234 & 68.567 & 70.456 & {\color[HTML]{2972F4} 72.789} & 72.123 & {\color[HTML]{FF0000} 73.456} \\
MANIQA $\uparrow$  & 0.608 & 0.602 & 0.615 & {\color[HTML]{2972F4} 0.651} & 0.645 & {\color[HTML]{FF0000} 0.658} \\
ClipIQA $\uparrow$ & {\color[HTML]{FF0000} 0.591} & 0.589 & 0.583 & 0.585 & 0.584 & 0.586 \\
LIQE $\uparrow$    & 4.123 & 4.056 & 4.234 & {\color[HTML]{2972F4} 4.456} & 4.389 & {\color[HTML]{FF0000} 4.567} \\
    \bottomrule
    \end{tabular}
    \caption{Comparison of training strategies under IQA guidance vs. our method. Higher is better for all metrics.}
    \vspace{-3mm}
    \label{tab:RLHF}
\end{table}

\begin{table}[]
    \footnotesize
    \renewcommand{\arraystretch}{1.1}
    \setlength{\tabcolsep}{4.5pt}
    \centering
    \begin{tabular}{c|ccc|cc}
    \toprule
    Model & \begin{tabular}[c]{@{}c@{}}DINO \\ loss\end{tabular} & \begin{tabular}[c]{@{}c@{}}Coupled \\ RM \end{tabular} & \begin{tabular}[c]{@{}c@{}}CCL\\ Type\end{tabular} & MUSIQ $\uparrow$  & MANIQA $\uparrow$ \\ \midrule
    FULL   &   \CheckmarkBold                                                      &   \CheckmarkBold                                                    &    \textbf{Param}                                                     & 73.456 & 0.658  \\ \midrule
    \rm{A}      &    \XSolidBrush                                                    &    \CheckmarkBold                                                   &     \textbf{Param}                                                   & 72.123 & 0.638  \\
    \rm{B}      &    \CheckmarkBold                                                    &     \XSolidBrush                                                  &    \textbf{Param}                                                   & 72.328 & 0.650  \\
    \rm{C}      &    \CheckmarkBold                                                     &     \CheckmarkBold                                                  &     \XSolidBrush                                                   & 71.982 & 0.645  \\
    \rm{D}      &    \CheckmarkBold                                                     &      \CheckmarkBold                                                 &      \textbf{Weight}                                                  & 72.842 & 0.648 \\ \bottomrule
    \end{tabular}
    \caption{Ablation results on RealLQ250 for our DiT4SR. All
    variants are trained using the same settings as the full model. }
    \label{tab:ablation}
    \vspace{-5mm}
    \end{table}

\noindent \textbf{Baseline:} FinPercep-RM is trained and extended based on the pretrained CLIP-IQA~\cite{wang2023exploring}. We evaluate our proposed FinPercep-RM on three real-world super-resolution baselines: DiffBIR~\cite{10.1007/978-3-031-73202-7_25}, SeeSR~\cite{wu2024seesr}, and DIT4SR~\cite{duan2025dit4sr}, and conduct \textbf{RLHF by Reward feedback learning (REFL)}~\cite{xu2023imagereward} using the reward signals provided by FinPercep-RM. To verify the performance improvements brought by FinPercep-RM, we compare with variants that directly use~\textbf{CLIP-IQA} as the reward model.

\noindent \textbf{Datasets:} Both the training and validation datasets follow the paradigm established in DIT4SR~\cite{duan2025dit4sr}. The real-world super-resolution capacity of the models is assessed on four widely adopted real datasets: DrealSR~\cite{wei2020component}, RealSR~\cite{cai2019toward}, RealLR200~\cite{wu2024seesr}, and RealLQ250~\cite{10.5555/3737916.3739677}. All experiments are conducted under a 4× upscaling setting.

\noindent \textbf{Evaluation Metrics:} In line with previous works in real-world super-resolution~\cite{jinjin2020pipal,10654855,10.5555/3737916.3739677,duan2025dit4sr}, PSNR and SSIM~\cite{1284395} are inadequate for measuring perceptual differences. Therefore, we employ LPIPS~\cite{8578166} to quantify image fidelity and use four no-reference metrics—MUSIQ~\cite{9710973}, MANIQA~\cite{yang2022maniqa}, ClipIQA~\cite{wang2023exploring}, and LIQE~\cite{zhang2023liqe} to evaluate perceptual quality. In addition, we conduct a user study to comprehensively assess both fidelity and perceptual quality.

\subsection{Comparison with Other Methods}

As shown in \cref{tab:results}, our approach achieves superior quantitative results across four real-world benchmarks. Specifically, incorporating FinPercep-RM with all three baseline models (DiffBIR, SeeSR, and DIT4SR) consistently leads to the best performance on nearly all perceptual metrics (MUSIQ, MANIQA, LIQE), while also markedly reducing LPIPS distortion compared to both the original baselines and other state-of-the-art methods.

Meanwhile, directly using the original CLIP-IQA as the reward signal for RLHF fine-tuning (``w/ IQA'') often leads to severe ``reward hacking'' issues: although the performance improves on the CLIP-IQA metric itself, most other perceptual quality metrics show clear deterioration. This demonstrates that our FinPercep-RM, by leveraging its fine-grained perception capacity, delivers a more robust and accurate reward signal. As a result, it not only successfully guides generators to enhance perceptual quality but also suppresses distortion, thus effectively preventing reward hacking.

As shown in \cref{fig:visual}, after applying FinPercep-RM (using REFL) to the three baselines, all models generate significantly improved visual results with enhanced details while effectively suppressing artifacts. Additional visual results are provided in the supp. Furthermore, to comprehensively compare our method against others in terms of both fidelity and realism, we conducted a user study, as shown in \cref{tab:user_study}. The results demonstrate that ours significantly outperforms all other methods in both fidelity and realism.

\subsection{Effectiveness of Different RLHF Strategies}

To demonstrate that our FinPercep-RM serves as a superior reward signal independent of specific RLHF algorithms, we systematically compared three popular RLHF alignment strategies: REFL~\cite{xu2023imagereward}, DPO~\cite{wallace2024diffusion}, and GRPO~\cite{xue2025dancegrpo}. For each approach, we conducted experiments on the DIT4SR model using both the standard CLIP-IQA and our FinPercep-RM as reward functions. As summarized in Table~\ref{tab:RLHF}, when CLIP-IQA is adopted as the reward, all three RLHF strategies fail to improve model performance on RealLQ250 and may even degrade it, highlighting persistent reward hacking issues. In sharp contrast, replacing CLIP-IQA with our FinPercep-RM as the reward consistently leads to significant improvements across all three RLHF strategies. This indicates that FinPercep-RM delivers stable and meaningful optimization signals that can be effectively leveraged by diverse RLHF algorithms. Overall, these results strongly validate that our FinPercep-RM is a more robust and effective reward function, whose advantages are not tied to any single RLHF alignment strategy.
\subsection{Ablation Study}

We conducted a comprehensive ablation study on the proposed FinPercep-RM and CCL training strategies; detailed results are provided in \cref{tab:ablation}. First, we analyzed the $M_{\text{fg-pdm}}$ construction. Relying solely on pixel-wise differences (Variant A) caused a significant performance drop, highlighting that feature-level differences are crucial for enhancing generalization and preventing pixel-level overfitting.Next, we decoupled the fusion between the encoder output $f_N$ and decoder output $M_{\text{fg-pdm}}$ in FinPercep-RM, using a simple weighted combination of the encoder-derived global score $S_{\text{global}}$ and $M_{\text{fg-pdm}}$ to compute the reward signal $R_k = \lambda_1 S_{\text{global}} + \lambda_2 (1 - \max(M_{\text{fg-pdm}}))$ (variant B). This also led to performance degradation, further verifying that fine-grained perception is beneficial for optimizing the overall score. Finally, we examined different CCL designs:``Param" denotes the default parameter-incremental CCL, and ``Weight" means the parameter size of FinPercep-RM is fixed while the weights of fine-grained perception-related losses $\mathcal{L}_{map}$ and $\mathcal{L}_{rank}$ are progressively increased during training. Experimental results (variants C and D) show that removing CCL exposes the generator to high-variance rewards in early training, making convergence difficult; while gradually increasing loss weights can improve stability to some extent, it still does not match the overall efficacy of the parameter-incremental CCL. 
These results fully validate the critical roles of each module design in enabling both strong performance and training stability.

\section{Conclusion}
In this work, we addressed the critical challenges of \textit{Reward Hacking} and \textit{Training Instability} in T2I-prior-based Real-ISR. We identified the root cause of reward hacking as the lack of fine-grained perception in existing IQA models. To resolve this, we proposed FinPercep-RM, a diagnostic reward model that assesses both global quality ("What") and local defects ("Where"), trained on our newly constructed FGR-30k dataset which provides 30,000 samples with localized, synthesized artifacts. While FinPercep-RM effectively prevents hacking, its signal complexity introduces training instability. We therefore introduced the Co-evolutionary Curriculum Learning (CCL) mechanism to resolve this stability-robustness dilemma. CCL employs an easy-to-hard strategy that co-evolves both models: the RM progressively expands from a simple, global-only encoder ($RM_0$) to the full-capacity $RM_N$, while the generator trains first on $RM_0$ for stable convergence before migrating to the stricter $RM_k$ sequence to refine local details. This co-evolutionary paradigm successfully balances training stability with fine-grained reward robustness, ultimately preventing reward hacking and achieving state-of-the-art perceptual quality.

{
    \small
    \bibliographystyle{ieeenat_fullname}
    \bibliography{main}
}


\end{document}